\documentclass[10pt,twocolumn,letterpaper]{article}

\usepackage{cvpr}
\usepackage{times}
\usepackage{epsfig}
\usepackage{graphicx}
\usepackage{amsmath}
\usepackage{amssymb}

\usepackage{amsthm}
\def\etal{\emph{et al}.}

\def\eg{\emph{e.g.}}
\def\ie{\emph{i.e.}}
\def\G{\mathcal{G}}
\def\R{\mathbb{R}}

\def\P{\mathcal{P}}
\def\T{\mathrm{T}}
\usepackage{multirow}
\usepackage{array}  
\usepackage[ruled,vlined]{algorithm2e}
\renewcommand{\KwResult}{\textbf{Output:}}
\renewcommand{\KwData}{\textbf{Input:}}

\usepackage[pagebackref=true,breaklinks=true,letterpaper=true,colorlinks,bookmarks=false]{hyperref}

\cvprfinalcopy 

\def\cvprPaperID{915} 

\ifcvprfinal\fi
\begin{document}

\title{Scalable Person Re-identification on Supervised Smoothed Manifold}

\author{Song Bai$^1$, Xiang Bai$^{1}$, Qi Tian$^2$\\
$^1$Huazhong University of Science and Technology,~~$^2$University of Texas at San Antonio\\
{\tt\small\{songbai,xbai\}@hust.edu.cn,~qi.tian@utsa.edu}
}

\maketitle

\begin{abstract}
Most existing person re-identification algorithms either extract robust visual features or learn discriminative metrics for person images. However, the underlying manifold which those images reside on is rarely investigated. That raises a problem that the learned metric is not smooth with respect to the local geometry structure of the data manifold.

In this paper, we study person re-identification with manifold-based affinity learning, which did not receive enough attention from this area. An unconventional manifold-preserving algorithm is proposed, which can 1) make the best use of supervision from training data, whose label information is given as pairwise constraints; 2) scale up to large repositories with low on-line time complexity; and 3) be plunged into most existing algorithms, serving as a generic postprocessing procedure to further boost the identification accuracies. Extensive experimental results on five popular person re-identification benchmarks consistently demonstrate the effectiveness of our method. Especially, on the largest CUHK03 and Market-1501, our method outperforms the state-of-the-art alternatives by a large margin with high efficiency, which is more appropriate for practical applications.
\end{abstract}

\section{Introduction}
\begin{figure*}[tb]
\centering
\includegraphics[width = 0.8\textwidth]{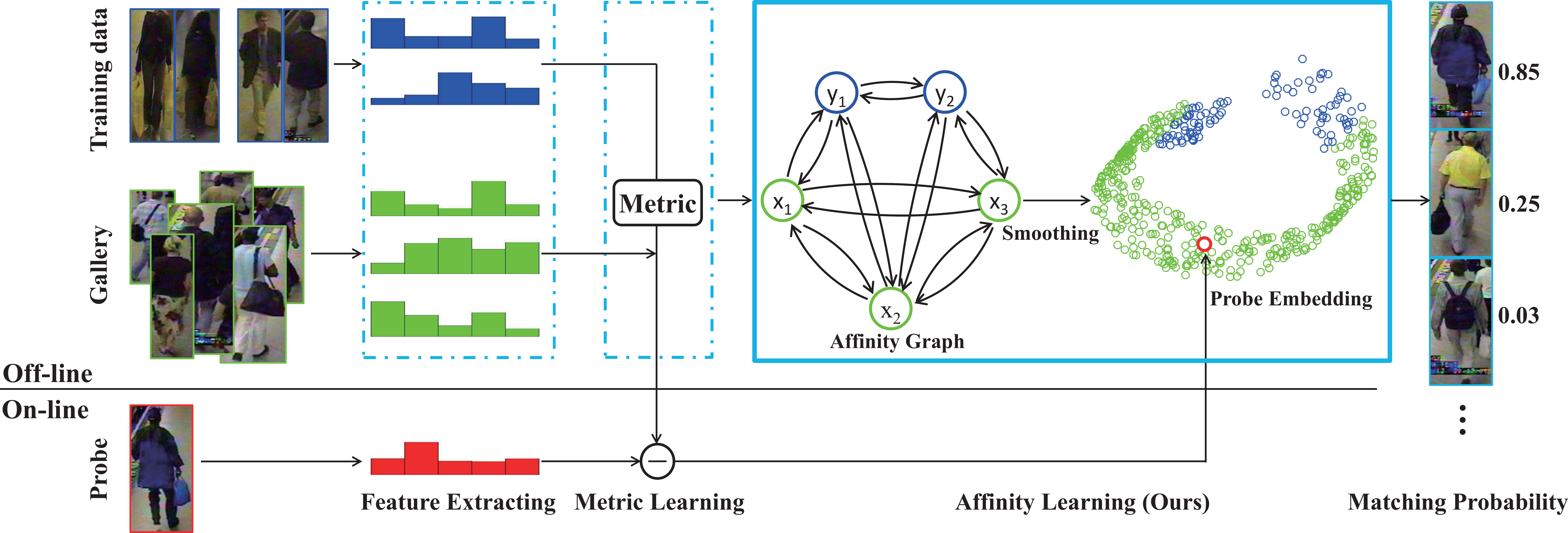}
\caption{The pipeline of a person re-identification system. The blue, green and red color indicate training data, gallery and probe, respectively. Previous works concentrate on feature extracting and metric learning, marked with dashed boxes. Our work can be the postprocessing procedure about affinity learning, marked with a solid box. Sample images come from GRID dataset~\cite{GRID1}.}
\label{fig:pipeline}
\vspace{-2ex}
\end{figure*}
Person re-identification (ReID) is an active task driven by the applications of visual surveillance, which aims to identify person images from the gallery that share the same identity as the given probe.
Due to the large intra-class variations in viewpoint, pose, illumination, blur and occlusion, person re-identification is still a rather challenging task, though extensively studied in recent years.

Current research interests can be coarsely divided into two mainstreams: 1) those focus on designing robust visual descriptors~\cite{SCNCD,symmetry,GOG,XQDA,zheng2015query} to accurately model the appearance of person; 2) those seek for a discriminative metric~\cite{PRDC,li2013decision,xiong2014person,lisanti2015person,hirzer2012relaxed}, under which instances of the same identity should be closer while instances of different identities are far away.

Unlike those methods performed in the metric space, we investigate person re-identification task from another perspective,~\ie,~taking into account the manifold structure~\cite{LLE}. Since existing methods only analyze the pairwise distances between instances, the underlying data manifold, which those images reside on, is more or less neglected.
It results in that the learned relationships (similarities or dissimilarities) between instances are not smooth with respect to the local geometry of the manifold.

To overcome this issue, potential solutions can be semi-supervised~\cite{LP,LGC} or unsupervised~\cite{zhou2004ranking,zhang2015query,donoser2013diffusion,RDP_AAAI} algorithms about manifold learning. However, directly applying such algorithms to person re-identification might be problematic for two reasons.
First, semi-supervised algorithms (\eg,~label propagation~\cite{LP}) can only predict the labels of unlabeled data, but fail to depict the relationship between the probe and gallery instances. Moreover, they require category labels, while supervision in ReID is given as pairwise (equivalence) constraints~\cite{KISSME}. Meanwhile, unsupervised algorithms (\eg,~manifold ranking~\cite{zhou2004ranking}, graph transduction~\cite{GT}) totally ignore the beneficial influence from the labeled training data.
Second, since most manifold learning algorithms operate on graph models, their algorithmic complexity is usually high. Therefore, the heavy computational cost hinders their promotions in this field, especially in recent years researchers begin to attach more importance to the scalability issue~\cite{Deepreid,market1501}. In summary, due to the above factors, those conventional manifold learning algorithms are inadequate to derive a more faithful similarity for person re-identification.

In this paper, we tackle person re-identification task on the data manifold by proposing a novel affinity learning algorithm called Supervised Smoothed Manifold (SSM).
Compared with existing algorithms, the primary contribution of SSM is that the similarity value between two instances is estimated in the context of other pairs of instances, thus the learned similarity well reflects the geometry structure of the underlying manifold.

Moreover, SSM is customized specifically for person re-identification, which further possesses three merits (as illustrated in Fig.~\ref{fig:pipeline}) as follows:
i)~\textbf{supervision}: instead of considering each instance individually, we propose to learn the similarity with instance pairs. By doing so, SSM can take advantage of the supervision in pairwise constraints, which is easily accessible in this task;
ii)~\textbf{efficiency}: to overcome the limitation of high time complexity of SSM, two improvements are proposed to accelerate its on-line person matching. Consequently, the affinity learning is performed only with database instances off-line, and SSM can be applied to the scenario on large scale person re-identification;
and iii)~\textbf{generalization}: different from most existing algorithms performed in metric space, SSM focuses on affinity learning between instances. Hence, SSM can be deemed as a postprocessing procedure (or a generic tool) to further boost the identification accuracies of those algorithms.

The rest of the paper is organized as follows. In Sec.~\ref{sec:R_W}, we present the differences between SSM and relevant works. The basic affinity learning framework of SSM is introduced in Sec.~\ref{sec:proposed}, and significantly accelerated in Sec.~\ref{sec:fly}. Experiments are presented in Sec.~\ref{sec:exp}. Conclusions and future works are given in Sec.~\ref{sec:con}.

\section{Related Work} \label{sec:R_W}
The manifold structure has been observed by several works. Motivated by the fact that pedestrian data are distributed on a highly curved manifold, a sampling strategy for training neural network called Moderate Positive Mining (MPM) is proposed in~\cite{shi2016embedding}. However, considering the data distribution is hard to define, MPM does aim at estimating the geodesic distances along the manifold. From this point of view, SSM explicitly learns the geodesic distances between instances, which can be directly used for re-identification.

Manifold ranking~\cite{zhou2004ranking} is introduced by~\cite{person_manifold_ranking} to person re-identification. Through a random walk~\cite{random_walk} on the affinity graph, it propagates the probe label to the gallery iteratively assuming that the probe is the only labeled data. Despite the ignorance of labeled training data as analyzed above, manifold ranking encounters severe obstacles when handling larger databases, since the graph-based iteration has to be run each time a new probe is observed.
In this aspect, SSM also learns the similarities via iterative propagation. Nevertheless, it enables a highly-efficient on-line matching.

Post-ranking techniques have not drawn much attention in this field. Most of them require human feedback in-the-loop~\cite{ali2010interactive,hirzer2011person}, such as Post-rank OPtimisation (POP)~\cite{POP}, Human Verification Incremental Learning (HVIL)~\cite{wang2016human}. Meanwhile, several works~\cite{an2016person,leng2015person} operate in an unsupervised manner. For example, Discriminant Context Information Analysis (DCIA)~\cite{DCIA} focuses on the visual ambiguities shared between the first ranks, where the true match is supposed to be located. In comparison, SSM does not need human interaction or hold the ``rank-1" assumption. Instead, its essence is to learn a smooth similarity measure, supervised by the special kind of labels in pairwise constraints.

At the first glance, affinity learning in our work appears the same as similarity learning (\eg,~PolyMap~\cite{PolyMap}). Unlike similarity learning on polynomial feature map~\cite{SCSP} which connects to Mahalanobis distance metric and bilinear similarity, affinity learning in SSM does not rely on the definition of metric (non-metric can be also used). Therefore, they are inherently different.
Finally, it is acknowledged that those metric learning methods (\eg,~KISSME~\cite{KISSME}, XQDA~\cite{XQDA}) are also relevant, but take effects prior to SSM in a person re-identification system as Fig.~\ref{fig:pipeline} shows.

\section{Proposed Method} \label{sec:proposed}
Given a probe $p$ and a testing gallery $X=\{x_1,x_2,\dots,x_{N_g}\}$, we aim at learning a smooth similarity $Q\in\R^{N\times N}$ with the help of the labeled training set $Y=\{y_1,y_2,\dots,y_{N_l}\}$, where $N=N_g+N_l+1$.
The data manifold is modeled as a weighted affinity graph $\G=\{V,W\}$. The vertex set $V=\{v_1,v_2,\dots,v_N\}$ is equivalent to the union of the probe $p$ and the database instances (gallery $X$ and labeled set $Y$). $W\in\R^{N\times N}$ is the adjacency matrix of $\G$, with $W_{ij}$ measuring the similarity between vertex $v_i$ and $v_j$.
To facilitate a random walk~\cite{random_walk} on the graph $\G$, a transition matrix $P\in\R^{N\times N}$ is usually needed. The transition probability from vertex $v_i$ to $v_j$ can be calculated as
\begin{equation}
P(i\rightarrow j)=P_{ij}=\frac{W_{ij}}{\sum_{j'=1}^NW_{ij'}}.
\end{equation}
Thus, $P$ is a row stochastic matrix.

\subsection{Supervised Similarity Propagation}
The label set $L\in\R^{N\times N}$ used in person re-identification is given in pairwise constraints,~\ie,~if $v_i$ and $v_j$ belong to the same identity, $L_{ij}=1$, otherwise $L_{ij}=0$. Meanwhile, in the ideal case, the learned similarity $Q_{ij}$ should be larger if $v_i$ and $v_j$ belong to the same identity, and $Q_{ij}$ should be close to $0$ otherwise. Therefore, we can conclude that both $L$ and $Q$ provide a probabilistic interpretation to the likelihood of the tuple $(v_i, v_j)$ being a true matching pair. The difference is that $L_{ij}$ is a discrete binary variable, indicating exactly matching or not, while $Q_{ij}$ is a continuous variable, specifying a matching degree. Such an observation motivates us that affinity learning can be done by propagating the pairwise constraint label $L$ with tuples as primitive data. In other words, similarities are spread from the most confident tuples generated from the labeled set $Y$ to the unexplored tuples generated from the testing gallery $X$.

Let $(v_k,v_i)$ and $(v_l,v_j)$ be two tuples, the propagation step in the $t$-th iteration is defined as
\begin{equation} \label{eq:iteration}
Q^{(t+1)}_{ki}=\alpha\sum_{l,j}^N\P(ki\rightarrow lj)Q^{(t)}_{lj}+(1-\alpha)L_{ki},
\end{equation}
where $\P(ki\rightarrow lj)$ is the transition probability from tuple $(v_k,v_i)$ to tuple $(v_l,v_j)$, and $0<\alpha<1$. Eq.~\eqref{eq:iteration} reveals that at each iteration, the tuple $(v_k,v_i)$ absorbs a fraction of label information from the rest tuples with probability $\alpha$, then retains its initial label $L_{ki}$ with probability $1-\alpha$.
Assuming the independence within tuples, we hold the \emph{product rule} to calculate $\P(ki\rightarrow lj)$, as
\begin{equation}
\P(ki\rightarrow lj)=P(k\rightarrow l)P(i\rightarrow j)=P_{kl}P_{ij}.
\end{equation}
Afterwards, Eq.~\eqref{eq:iteration} can be rewritten in matrix form
\begin{equation} \label{eq:iteration1}
\vec{Q}^{(t+1)}=\alpha\P\vec{Q}^{(t)}+(1-\alpha)\vec{L}.
\end{equation}

To prove this, we need two identical coordinate transformations, that is $\mu\equiv N(i-1)+k$ and $\nu\equiv N(j-1)+l$.
Then $Q$ can be vectorized to $\vec{Q}=vec(Q)\in\R^{N^2\times 1}$, with the element correspondence $\vec{Q}_{\mu}=Q_{ki}$. Let $\P\in\R^{N^2\times N^2}$ be the Kronecker product of $P$ with itself,~\ie,~$\P=P\otimes P$. Then, the correspondence between $\P$ and $P$ is given as $\P_{\mu\nu}=P_{ij}P_{kl}$. Eventually, Eq.~\eqref{eq:iteration} can be expressed as
\begin{equation}
\vec{Q}^{(t+1)}_{\mu}=\alpha\sum_{\nu=1}^{N^2}\P_{\mu\nu}\vec{Q}^{(t)}_{\nu}+(1-\alpha)\vec{L}_{\mu}.
\end{equation}
The proof is complete.

\subsection{Convergence Proof}
By running the iteration for $t$ times, Eq.~\eqref{eq:iteration1} can be expanded as
\begin{equation} \label{eq:iteration2}
\vec{Q}^{(t+1)}=({\alpha\P})^t\vec{Q}^{(1)}+(1-\alpha){\sum_{i=0}^{t-1}({\alpha\P})^i\vec{L}}.
\end{equation}
$\P$ is also a row stochastic matrix, since
\begin{equation}
\sum_{\nu}\P_{\mu\nu}=\sum_{l,j}P_{ij}P_{kl}=\sum_{j}{P_{ij}}\sum_l{P_{kl}}=1.
\end{equation}
Therefore, according to \emph{Perron-Frobenius Theorem}, we can obtain that spectral radius of $\P$ is bounded by $1$, the maximum value of its row sums. Considering that $0<\alpha<1$, we have
\begin{equation}
\lim_{t\rightarrow\infty}({\alpha\P})^t=0,~~~~\lim_{t\rightarrow\infty}{\sum_{i=0}^{t-1}({\alpha\P})^i}=(I-\alpha\P)^{-1},
\end{equation}
where $I$ is an identity matrix in appropriate size.
Consequently, Eq.~\eqref{eq:iteration2} converges to
\begin{equation} \label{eq:close}
\lim_{t\rightarrow\infty}\vec{Q}^{(t+1)}=(1-\alpha)(I-\alpha\P)^{-1}\vec{L}.
\end{equation}
Then $Q$ can be obtained by reshaping $\vec{Q}$ to matrix form as $Q=vec^{-1}(\vec{Q})$.

\subsection{Basic Pipeline}
Intuitively, person re-identification using the above affinity learning algorithm can be accomplished in three steps.
First, each time a probe instance $p$ is observed, the affinity graph $\G$ is constructed.
Second, a new similarity $Q$ is learned by either running Eq.~\eqref{eq:iteration1} until convergence or directly using the closed-form solution in Eq.~\eqref{eq:close}.
At last, since $Q$ can be divided into
\begin{equation}
Q=
 \begin{bmatrix}
   Q_{pp} & Q_{pX} & Q_{pY} \\
   Q_{Xp} & Q_{XX} & Q_{XY} \\
   Q_{Yp} & Q_{YX} & Q_{YY} \\
  \end{bmatrix},
\end{equation}
we can obtain the matching probabilities between the probe $p$ and the gallery $X$, that is $Q_{pX}\in\R^{1\times N_g}$. Note that $W$ and $P$ also have such a division.

We draw readers' attention that when the probe $p$ is used for testing, the other probe instances are invisible to users. Therefore, one cannot simultaneously include all the probe instances to constitute $\G$ for a global probe search.

However, this pipeline is computationally too demanding in practice.
First, affinity learning itself is computationally expensive. It requires time complexity $O(TN^4)$ and space complexity $O(N^4)$ to run the iteration in Eq.~\eqref{eq:iteration1}, where $T$ is the iteration number. Alternatively, using the closed-form solution in Eq.~\eqref{eq:close} requires time complexity $O(N^6)$ and space complexity $O(N^4)$, since we need to invert and store a huge matrix of size $N^2\times N^2$.

Second, adapting new probe instances is computationally expensive. As our method is algorithmically graph-based, we need to discard the old probe and do the affinity learning at each time a new probe is observed.
Assume we have $N_p$ probe instances in total, we at least need time complexity $O(TN_pN^4)$ to finish the whole probe search. Note that constructing the affinity graph is computationally cheap due to the fact that the similarities between database instances can be pre-computed off-line for once and reused consistently.

\section{Re-identification on-the-fly} \label{sec:fly}
In this section, we propose two modifications to decrease the high complexity of the basic pipeline in Sec.~\ref{sec:proposed}, such that person re-identification can be done on-the-fly.

\subsection{Iteration Transform}
Our first improvement focuses on affinity learning itself. We observe the following useful identity
\begin{equation}
\P\vec{Q}=(P\otimes P) vec(Q)=vec(PQP^{\T}).
\end{equation}
So, Eq.~\eqref{eq:iteration1} can be transformed into
\begin{equation} \label{eq:iteration_effeiciency}
Q^{(t+1)}=\alpha PQ^{(t)}P^{\T}+(1-\alpha)L.
\end{equation}
As a result, the time and space complexity of affinity learning are reduced to $O(TN^3)$ and $O(N^2)$, respectively.

\subsection{Probe Embedding}
Our second improvement concentrates on improving the efficiency in adapting new probe instances. First, we prove that the closed-form solution in Eq.~\eqref{eq:close} can be derived from
\begin{equation} \label{eq:regularization}
\min_{Q}\Phi(Q)+\frac{1-\alpha}\alpha\Omega(Q),
\end{equation}
where
\begin{equation} \label{eq:twoQ}
\begin{split}
&\Phi(Q)=\frac12\sum_{i,j,k,l}^N{P_{ij}P_{kl}}(Q_{ki}-Q_{lj})^2,\\
&\Omega(Q)=\sum_{k,i=1}^N(Q_{ki}-L_{ki})^2.
\end{split}
\end{equation}
Using the two identical coordinate transformations, Eq.~\eqref{eq:regularization} can be vectorized, where
\begin{equation} \label{eq:regularization1}
\Phi(\vec{Q})=\frac12\sum_{\mu,\nu}^{N^2}\P_{\mu\nu}(\vec{Q}_{\mu}-\vec{Q}_{\nu})^2,~~\Omega(\vec{Q})=\|\vec{Q}-\vec{L}\|_2^2,
\end{equation}
where $\Phi(\vec{Q})$ measures the smoothness of $\vec{Q}$ with respect to the local manifold structure, and $\Omega(\vec{Q})$ measures the fitness of $\vec{Q}$ to the given label $\vec{L}$.

The derivative of $\Phi(\vec{Q})$ with respect to $\vec{Q}$ is
\begin{equation} \label{eq:Phi2Q1}
\frac{\partial\Phi(\vec{Q})}{\partial{\vec{Q}}}=\left((I-\P)+(I-\P)^{\T}\right)\vec{Q}.
\end{equation}
According to~\cite{belkin2003laplacian}, Eq.~\eqref{eq:Phi2Q1} can be approximated by
$2(I-\P)\vec{Q}$.
So, one can easily induce the derivative of Eq.~\eqref{eq:regularization} with respect to $\vec{Q}$
\begin{equation} \label{eq:derivative}
2(I-\P)\vec{Q}+\frac{2(1-\alpha)}\alpha(\vec{Q}-\vec{L}).
\end{equation}
By setting Eq.~\eqref{eq:derivative} to zero and applying $vec^{-1}$ operator, we can get the closed-form solution of Eq.~\eqref{eq:regularization}
\begin{equation}
Q=vec^{-1}\left((1-\alpha)(I-\alpha\P)^{-1}\vec{L}\right),
\end{equation}
which is equivalent to Eq.~\eqref{eq:close}. The proof is complete.

Compared with the large database (testing gallery and labeled data), there is only one probe $p$ at each testing time. Therefore, we hold two assumptions that 1) the database itself constitutes an underlying manifold; 2) when $p$ is embedded into the manifold smoothly, it will not alter its geometry structure. With these prerequisites, we can first perform affinity learning off-line with only database instances, then do the probe embedding on-line.

Of course, the embedding of the probe should also follow the smoothness criterion $\Phi(Q)$.
After the pairwise similarities between database instances are smoothed, the partial derivative of $\Phi(Q)$ with respect to $Q_{pi}$ is
\begin{equation}
\frac{\partial \Phi(Q)}{\partial{Q_{pi}}}=\sum_{j,l=1}^{N_g+N_l}P_{ij}P_{pl}(Q_{pi}-Q_{lj}).
\end{equation}
Setting it to zero, the similarity between the probe $p$ and a certain database instance $v_i$ can be calculated
\begin{equation} \label{eq:embedding_solution}
Q_{pi}=\sum_{j,l=1}^{N_g+N_l}P_{pl}Q_{lj}P_{ij}.
\end{equation}
By varying $v_i\in X$, Eq.~\eqref{eq:embedding_solution} can be rewritten in matrix form
\begin{equation} \label{eq:embedding_solution_matrix}
Q_{pX}=\begin{bmatrix}P_{pX}&P_{pY}\end{bmatrix}\begin{bmatrix}Q_{XX}&Q_{XY}\\Q_{YX}&Q_{YY}\end{bmatrix}\begin{bmatrix}P_{XX}^{\T}\\P_{XY}^{\T}\end{bmatrix}.
\end{equation}

\subsection{Complexity Analysis} \label{sec:complexity}
The final pipeline of the proposed SSM is rather simple, summarized in Alg.~\ref{alg}.
\begin{algorithm}[tb]
\DontPrintSemicolon
\KwData{~The probe $p$, the testing gallery $X$, the labeled data $Y$, the training label $L$.\\}
\KwResult{~The matching probability $Q_{pX}$. \\}
\Begin{\emph{Off-line:} \\
    \Begin{
    Construct the affinity graph with $X$ and $Y$;\\
    Affinity learning with label $L$ using Eq.~\eqref{eq:iteration_effeiciency}. \\
    \KwRet{$Q_{XX}$, $Q_{XY}$, $Q_{YX}$, $Q_{YY}$}
    }
    \emph{On-line:} \\
    \Begin{
    \For{each probe $p$}{
    Do pedestrian matching using Eq.~\eqref{eq:embedding_solution_matrix};\\
    \KwRet{$Q_{pX}$}.}
    }
}
\caption{Supervised Smoothed Manifold.\label{alg}}
\end{algorithm}
As can be seen, affinity learning is done only with database instances.
The computational cost still seems a bit heavy, since there are $(N_g+N_l)$ vertices in the graph. However, those operations can be done off-line, and reused with different probe instances. The learned similarities can all be maintained dynamically as long as new database instances are added or distance matrices are changed.

In Table~\ref{table:complexity}, we present the on-line complexity comparison between the standard solution in Sec.~\ref{sec:proposed} and the accelerated solution in Sec.~\ref{sec:fly}. Eq.~\eqref{eq:embedding_solution_matrix} reveals that on-line indexing for $N_p$ probe instances involves the multiplication of three matrices. Whereas the multiplication of the right two can be also computed off-line, the on-line time complexity is only $O\left (N_p(N_g+N_l)N_g\right)$. Furthermore, the space complexity is dominated by the storage of the learned similarity, requiring $O\left((N_g+N_l)^2\right)$.
\begin{table}[tb]
\small
\centering
\begin{tabular}{|l|*{2}{p{2.75cm}<{\centering}}|}
\hline
Methods & Time Complexity & Space Complexity \\
\hline
\hline
Standard & $O(TN_pN^4)$ & $O(N^4)$ \\
Accelerated & $O\left(N_p(N_g+N_l)N_g\right)$ & $O\left((N_g+N_l)^2\right)$ \\
\hline
\end{tabular}
\caption{The complexity comparison between the standard solution and the accelerated solution of SSM. Recall that $N_p$ is the number of probe, $N_g$ is the number of gallery, $N_l$ is the number of labeled data, and $T$ is the number of iterations. $N=N_g+N_l+1$.}
\label{table:complexity}
\vspace{-2ex}
\end{table}

\section{Experiments} \label{sec:exp}
The proposed Supervised Smoothed Manifold (SSM) is evaluated on five popular benchmarks, including GRID~\cite{GRID1,GRID2}, VIPeR~\cite{VIPeR_ELF}, PRID450S~\cite{PRID450S}, CUHK03~\cite{Deepreid} and Market-1501~\cite{market1501}.
In the implementations of SSM, we do not carefully tune parameters, but fix $\alpha=0.1$ and the number of iterations $T=30$ throughout our experiments. The affinity graph is constructed by applying self-tuning~\cite{self-tuning} Gaussian kernel to pairwise distances following~\cite{person_manifold_ranking}.

\subsection{QMUL GRID}
QMUL underGround Re-IDentification (GRID)~\cite{GRID1,GRID2} is a challenging dataset, which has gradually become popular. The variations in the pose, colors and illuminations of pedestrians, as well as the poor image quality, make it very difficult to yield high matching accuracies.

GRID dataset consists of $250$ identities, with each identity having two images seen from different camera views. Besides, $775$ additional images that do not belong to the $250$ identities are used to enlarge the gallery. Sample images can be found in Fig.~\ref{fig:pipeline}. A fixed training/testing split with $10$ trials is provided. For each trial, $125$ image pairs are used for training. The remaining $125$ image pairs and the $775$ background images are used for testing. To evaluate the performances, we employ Cumulated Matching Characteristics (CMC) curves and the cumulated matching accuracy at selected ranks.

To obtain the image representations, we utilize two representative descriptors,~\ie,~Local Maximal Occurrence (LOMO)~\cite{XQDA} and Gaussian Of Gaussian (GOG)~\cite{GOG}. In addition, ELF6 feature~\cite{ELF6}, provided along with the dataset, is also tested to ensure the fair comparison.

\vspace{1ex}\noindent\textbf{Comparison with Baselines.}~In Table~\ref{table:baseline_GRID900}, we present the performances before and after SSM is used. Besides the three individual visual features, two types of fused features are also used. \emph{Fusion} means the concatenation of LOMO and GOG, while \emph{Fusion$^\star$} means the concatenation of all the three features, both with equal weights.
The pairwise distances between instances are computed in metric space. In our experiments, besides the natural choice of Euclidean metric, we also evaluate Cross-view Quadratic Discriminant Analysis (XQDA)~\cite{XQDA} which is taken as a representative of metric learning techniques.
\begin{table}[tb]
\small
\centering
\begin{tabular}{|l|*{2}{p{0.9cm}<{\centering}}|*{3}{p{0.95cm}<{\centering}}|}
\hline
Feature          & Metric  & Affinity & r=1 & r=10 & r=20      \\
\hline
\hline
ELF6 & Euclidean & $\times$ & 4.64 & 19.60 & 28.00 \\
ELF6 & Euclidean & $\surd$ & 6.96 & 23.44 & 34.08 \\
ELF6 & XQDA & $\times$ & 10.48 & 38.64 & \textbf{52.56} \\
ELF6 & XQDA & $\surd$  & \textbf{11.04}  & \textbf{40.72}  & 51.76 \\
\hline
\hline
LOMO & Euclidean & $\times$ & 15.20 & 30.80 & 36.40 \\
LOMO & Euclidean & $\surd$ & 16.00 & 33.68 & 41.60 \\
LOMO & XQDA & $\times$ & 16.56 & 41.84 & 52.40 \\
LOMO & XQDA & $\surd$ & \textbf{18.96} & \textbf{44.16} & \textbf{55.92} \\
\hline
\hline
GOG & Euclidean & $\times$ & 13.28 & 33.76 & 44.40 \\
GOG & Euclidean & $\surd$ & 14.40 & 36.80 & 44.48 \\
GOG  & XQDA & $\times$ & 24.80 & 58.40 & 68.88 \\
GOG  & XQDA & $\surd$ & \textbf{26.16} & \textbf{59.20} & \textbf{70.40} \\
\hline
\hline
Fusion & Euclidean & $\times$ & 14.72  & 35.44 & 45.84  \\
Fusion & Euclidean & $\surd$ & 17.76  & 37.60  & 44.48 \\
Fusion & XQDA & $\times$ & 27.04 & 59.36 & 70.00  \\
Fusion & XQDA & $\surd$ & \textbf{27.20} & \textbf{61.12} & \textbf{70.56} \\
\hline
\hline
Fusion$^\star$ & Euclidean & $\times$ & 14.80 & 35.60 & 46.24\\
Fusion$^\star$ & Euclidean & $\surd$ & 15.92 & 35.60 & 46.40\\
Fusion$^\star$ & XQDA & $\times$ & 27.20  & 61.12 & 71.20 \\
Fusion$^\star$ & XQDA & $\surd$ & \textbf{27.60} & \textbf{62.56} & \textbf{71.60} \\
\hline
\end{tabular}
\caption{The comparison with baselines on GRID dataset. $\surd$ indicates SSM is used and $\times$ indicates not used.}
\label{table:baseline_GRID900}
\vspace{-2ex}
\end{table}

As can be drawn, SSM leads to considerable performance gains against the baselines. For example, with ELF6 in Euclidean metric, the improvement of identification rate brought by SSM is $2.32$ at rank-$1$, $3.84$ at rank-$10$ and $6.08$ at rank-$10$. Meanwhile, by integrating XQDA, SSM can still boost the performances further. For example, the rank-1 accuracy of LOMO with XQDA is originally $16.56$, then increased to $18.96$ after the proposed SSM is used.
Those experimental results suggest that most existing visual features or metric learning algorithms in person re-identification are compatible with SSM. In other words, after visual features are given, person re-identification systems can be improved with two steps,~\ie,~applying metric learning first, and applying SSM next.

\begin{figure}[tb]
\centering
\includegraphics[width = 0.35\textwidth]{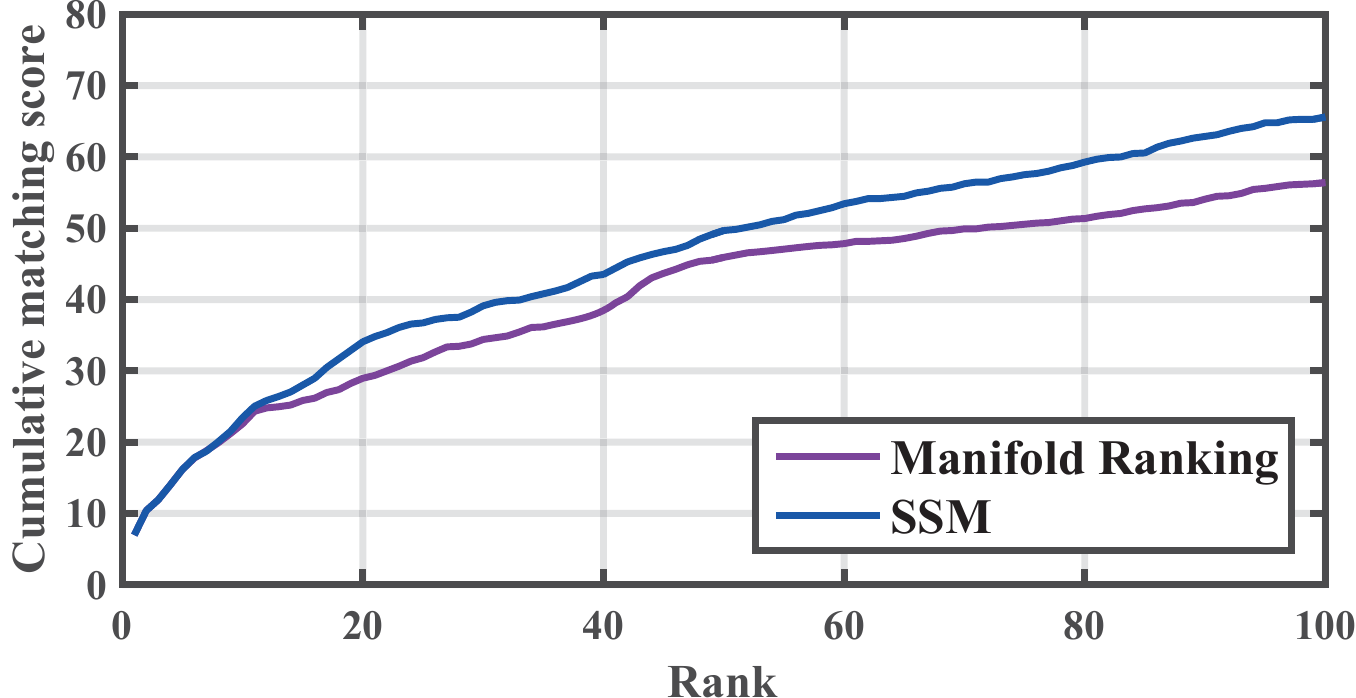}
\caption{The comparison between the proposed SSM and Manifold Ranking with the same affinity graph.}
\label{fig:ELF6}
\vspace{-2ex}
\end{figure}

As a related work to ours, manifold ranking~\cite{person_manifold_ranking} reports an identification rate of $30.96$ at rank-20 using ELF6 and Euclidean metric, which is significantly lower than $34.08$ achieved by SSM. It clearly demonstrates that it is beneficial to exploit the supervision information in affinity learning step. To avoid the performance uncertainty (though rather tiny) led by different implementation details, we compare SSM with manifold ranking using exactly the same affinity graph. The results are given in Fig.~\ref{fig:ELF6}. As we can see, the rank-1 accuracy of both manifold ranking and SSM is $6.96$. However, SSM outperforms manifold ranking by a large margin at higher ranks. The difference of accuracy is nearly $10$ at rank-100.

One of the most important properties of SSM is its high time efficiency in on-line pedestrian matching. Here, we omit the overhead in constructing the affinity graph, which can be done off-line. Under the same computing platform, manifold ranking takes $14.40$ seconds in total to fulfill searching $125$ probe instances, while SSM only needs $9.52ms$. One can easily find that SSM is $3$ orders of magnitude faster than manifold ranking. The reason behind is that the iteration of manifold ranking is conducted each time a new probe is observed. In comparison, SSM proposes to do affinity learning only off-line, and embed the probe smoothly into the manifold. As a result, although SSM has to do affinity learning on a much larger graph (to leverage the supervision from training data), the on-line cost can be controlled so that SSM has the potential ability of handling large-scale person re-identification. We will further discuss this aspect below.

From Table~\ref{table:baseline_GRID900}, failure cases of SSM can be also observed. As it suggests, the rank-20 accuracy of ELF6 under XQDA metric is originally $52.56$, then is decreased slightly by SSM to $51.76$. The  reason behind such abnormal phenomena is that the principle of SSM is to obtain a global similarity measure between each two instances, which varies smoothly with respect to the local geometry of the underlying manifold. The learned similarity cannot guarantee that the identification rate at specifical ranks will be improved. But in general cases, the overall performances will be refined.

\vspace{1ex}\noindent\textbf{Comparison with State-of-the-art.}~In Table~\ref{table:art_GRID900}, we give a thorough comparison with other state-of-the-art methods.
The performances of the proposed SSM are reported by using \emph{Fusion} feature (the concatenation of LOMO and GOG) under XQDA metric, which is a default configuration used in our later experiments.

Previous state-of-the-art performances are achieved by Spatially Constrained Similarity function on Polynomial feature map (SCSP)~\cite{SCSP} and GOG~\cite{GOG}. Chen~\etal~\cite{SCSP} impose spatially constraints to the similarity learning on polynomial feature map~\cite{PolyMap}, and report rank-1 accuracy $24.24$ by fusing $6$ visual cues. GOG~\cite{GOG} is a powerful descriptor proposed recently, which captures the mean and the covariance information of pixel features. With XQDA metric, it reports the best performances on GRID dataset,~\ie,~rank-1 accuracy $24.80$. Benefiting from~\emph{Fusion} feature and XQDA metric, SSM easily sets a new state-of-the-art performance, outperforming the previous by $2.40$ in rank-1 accuracy.

We emphasize that SSM is not restricted by the used descriptor and metric. Table~\ref{table:baseline_GRID900} presents that SSM can achieve higher performances with \emph{Fusion$^\star$} feature.
\begin{table}[tb]
\small
\centering
\begin{tabular}{|l|*{3}{p{1cm}<{\centering}}|}
\hline
Methods          & r=1 & r=10 & r=20      \\
\hline
\hline
ELF6+RankSVM~\cite{RankSVM}    & 10.24 & 33.28 & 43.68 \\
ELF6+PRDC~\cite{PRDC} & 9.68 & 32.96 & 44.32 \\
ELF6+RankSVM+MR~\cite{person_manifold_ranking} & 12.24 & 36.32 & 46.56 \\
ELF6+PRDC+MR~\cite{person_manifold_ranking} & 10.88 & 35.84 & 46.40 \\
ELF6 + XQDA~\cite{XQDA}  & 10.48 & 38.64 & 52.56 \\
LOMO + XQDA~\cite{XQDA}      & 16.56 &  41.84 &  52.40 \\
MLAPG~\cite{MLAPG} & 16.64 & 41.20 & 52.96 \\
NLML~\cite{NLML} & 24.54 & 43.53 & 55.25 \\
PolyMap~\cite{PolyMap}   & 16.30	 &	46.00	 & 	57.60 \\
SSDAL~\cite{SuChi2} & 22.40	& 48.00 &	58.40 \\
MtMCML~\cite{MtMCML} & 14.08	& 45.84	& 59.84 \\
LSSCDL~\cite{LSSCDL} & 22.40 & 51.28  & 61.20 \\
KEPLER~\cite{KEPLER} & 18.40	& 50.24	& 61.44 \\
DR-KISS~\cite{DR-KISS}	 & 20.60 & 51.40 & 62.60 \\
SCSP~\cite{SCSP} & 24.24	&	54.08	&	65.20 \\
GOG+XQDA~\cite{GOG} & \textbf{\color{blue}{24.80}} & \textbf{\color{blue}{58.40}} & \textbf{\color{blue}{68.88}} \\
\hline
\hline
SSM (Ours) & \textbf{\color{red}{27.20}} & \textbf{\color{red}{61.12}} & \textbf{\color{red}{70.56}} \\
\hline
\end{tabular}
\caption{The comparison with state-of-the-art on GRID dataset. The best and second best performances are marked in red and blue, respectively.}
\label{table:art_GRID900}
\vspace{-2ex}
\end{table}

\subsection{VIPeR, PRID450S and CUHK03}
VIPeR~\cite{VIPeR_ELF} is a widely-accepted benchmark for person re-identification containing $632$ identities, and PRID450S~\cite{PRID450S} consists of $450$ identities, both captured by two disjoint cameras. The widely adopted experimental protocol on two datasets is that a random selection of half persons is used for training and the rest for testing. The procedure is repeated for $10$ times, then the average performances are reported.

CUHK03~\cite{Deepreid} is among the largest public available benchmarks nowadays. It includes $13,164$ images of $1,360$ persons, with each person having $4.8$ images on average. Besides manually cropped images, auto detected images are also provided. Following the conventional experimental setup~\cite{Deepreid,XQDA,MLAPG,GOG,null}, $1,160$ persons are used for training and $100$ persons are used for testing. The experiments are conducted in single-shot setting with $20$ random splits.

In Table~\ref{table:VIPeR_PRID450s_CUHK03}, we present the performances of SSM and the baselines, where distances are calculated under XQDA metric. Consistent to previous experiments, SSM can easily boost the performances of baselines by around $2.5$ percent on average. In particular, the performance improvements are more dramatic on CUHK03. For example, the rank-1 accuracy of \emph{Fusion} is increased by $4.76$ on CUHK03 labeled dataset, and by $4.65$ on CUHK03 detected dataset. The preference of SSM on larger datasets stems from the fact that the manifold structure can be better sampled given more data points.
\begin{table*}[tb]
\small
\centering
\begin{tabular}{|l|*{3}{p{0.8cm}<{\centering}}|*{3}{p{0.8cm}<{\centering}}|*{3}{p{0.8cm}<{\centering}}|*{3}{p{0.8cm}<{\centering}}|}
\hline
\multirow{2}{*}{Methods} & \multicolumn{3}{c|}{VIPeR} & \multicolumn{3}{c|}{PRID450S} & \multicolumn{3}{c|}{CUHK03 (labeled)} & \multicolumn{3}{c|}{CUHK03 (detected)}   \\
\cline{2-13}
            & r=1 & r=10 & r=20 & r=1 & r=10 & r=20   & r=1 & r=5 & r=10  & r=1 & r=5 & r=10      \\
\hline
\hline
LOMO        & 40.00 & 80.51 & 91.08 & 61.38 & 91.02 & 95.33 & 50.85 & 81.38 & 91.14 & 44.45 & 78.70 & 87.65  \\
LOMO++SSM   & 42.22 & 83.54 & 92.82 & 62.84 & 92.62 & 96.49 & 52.50 & 84.53 & 92.49 & 49.05 & 81.25 & 90.30 \\
GOG         & 49.72 & 88.67 & 94.53 & 68.00 & 94.36 & 97.64 & 68.47 & 90.69 & 95.84 & 64.10 & 88.40 & 94.30 \\
GOG+SSM     & 50.73 & 89.97 & 95.63 & 68.49 & 95.73 & 98.53 & 71.82 & 92.54 & 96.64 & 68.20 & 90.30 & 94.10 \\
Fusion      & 53.26 & 90.95 & 95.73 & 72.04 & 95.82 & 98.49 & 71.87 & 92.64 & 96.80 & 68.05 & 90.15 & 94.95 \\
Fusion+SSM  & \textbf{53.73} & \textbf{91.49} & \textbf{96.08} & \textbf{72.98} & \textbf{96.76} & \textbf{99.11} & \textbf{76.63} & \textbf{94.59} & \textbf{97.95} & \textbf{72.70} & \textbf{92.40} & \textbf{96.05}\\
\hline
\end{tabular}
\caption{The comparison with baselines on VIPeR, PRID450S and CUHK03 dataset.}
\label{table:VIPeR_PRID450s_CUHK03}
\vspace{-2ex}
\end{table*}

\vspace{1ex}\noindent\textbf{Comparison on VIPeR.}~Since enormous algorithms have reported results on VIPeR dataset, it is less realistic to exhibit all of them. Hence, we only include those published in recent $3$ years or have close relationships with our work.
\begin{table}[tb]
\small
\centering
\begin{tabular}{|lc|*{3}{p{0.65cm}<{\centering}}|}
\hline
Methods        & Ref & r=1 & r=10 & r=20 \\
\hline
\hline
Local Fisher~\cite{local_fisher} & CVPR2013 & 24.18 & 67.12 & - \\
eSDC~\cite{eSDC} & CVPR2013 & 26.74 & 62.37 & 76.36 \\
SalMatch~\cite{SalMatch} & ICCV2013 & 30.16 & - & - \\
Mid-Filter~\cite{Mid-Filter} & CVPR2014 & 29.11 &  65.95 &  79.87 \\
SCNCD~\cite{SCNCD} & ECCV2014  & 37.80 & 81.20 & 90.40 \\
\hline
\hline
ImprovedDeep~\cite{ImprovedDeep} & CVPR2015 & 34.81 & - & - \\
PolyMap~\cite{PolyMap}  & CVPR2015  & 36.80 &	83.70 &	91.70 \\
XQDA~\cite{XQDA} & CVPR2015 & 40.00	& 80.51	& 91.08 \\
Semantic~\cite{Semantic} & CVPR2015 & 41.60 &	86.20 &	95.10\\
MetricEmsemb.~\cite{MetricEmsemble} & CVPR2015 & 45.90 & 88.90 & 95.80 \\
QALF~\cite{zheng2015query} & ICCV2015 & 30.17 & 62.44 & 73.81 \\
CSL~\cite{CSL} & ICCV2015 & 34.80 &	82.30 &	91.80 \\
MLAPG~\cite{MLAPG} & ICCV2015 & 40.73 &	82.34 & 92.37\\
MTL-LORAE~\cite{SuChi1} & ICCV2015 & 42.30 & 81.60 & 89.60 \\
DCIA~\cite{DCIA} & ICCV2015 & \textbf{\color{red}{63.92}} & 87.50 & - \\
\hline
\hline
DGD~\cite{DGD} & CVPR2016 & 38.60 & - & - \\
LSSCDL~\cite{LSSCDL} & CVPR2016 & 42.66 & 84.27 & 91.93 \\
TPC~\cite{TPC} & CVPR2016 & 47.80 & 84.80  & 91.10 \\
GOG~\cite{GOG} & CVPR2016 & 49.72 & 88.67 & 94.53 \\
Null~\cite{null} & CVPR2016 & 51.17	& \textbf{\color{blue}{90.51}}	& 95.92 \\
SCSP~\cite{SCSP} & CVPR2016 & 53.54	&  \textbf{\color{red}{91.49}} & \textbf{\color{red}{96.65}} \\
S-CNN~\cite{S-CNN} & ECCV2016 & 37.80 & 66.90 & - \\
Shi~\etal~\cite{shi2016embedding} & ECCV2016 & 40.91 & - & - \\
$\ell$1-graph~\cite{L1_graph} & ECCV2016 & 41.50 & - & - \\
S-LSTM~\cite{S-LSTM} & ECCV2016 & 42.40 & 79.40 &  - \\
SSDAL~\cite{SuChi2} & ECCV2016 & 43.50 & 81.50 & 89.00\\
TMA~\cite{TMA} & ECCV2016  & 48.19	& 87.65 & 93.54 \\
\hline
\hline
SSM (Ours) &  & \textbf{\color{blue}{53.73}} & \textbf{\color{red}{91.49}} & \textbf{\color{blue}{96.08}} \\
\hline
\end{tabular}
\caption{The comparison with state-of-the-art on VIPeR dataset.}
\label{table:VIPeR_art}
\vspace{-1ex}
\end{table}

The comparison is given in Table~\ref{table:VIPeR_art}. As can be seen, SSM yields the best rank-10 accuracy $91.49$, which is the same as SCSP~\cite{SCSP}. Meanwhile, SSM also achieves the second best performances at rank-1 and rank-20.
To our best knowledge now, the best rank-1 accuracy is achieved by Discriminant Context Information Analysis (DCIA)~\cite{DCIA}. The superiority of DCIA at rank-1 lies in that it tries to remove the visual ambiguities between the probe and its true match, which is supposed to be located at the first rank. By contrast, SSM does not hold such assumptions, which seem to be a bit strict in realistic settings. Thus, one can also observe that SSM outperforms DCIA by $3.99$ at rank-10. Considering their inherent difference of  principles, it can be anticipated that SSM and DCIA can benefit from each other, and a proper ensemble of them can lead to better performances.

\vspace{1ex}\noindent\textbf{Comparison on PRID450S.}~On PRID450S dataset, SSM provides the state-of-the-art performances on all the three evaluation metrics,~\ie,~$72.98$ at rank-1, $96.76$ at rank-10, and $99.11$ at rank-20. 
\begin{table}[tb]
\small
\centering
\begin{tabular}{|p{2cm}p{1.6cm}<{\centering}|*{3}{p{0.85cm}<{\centering}}|}
\hline
Methods        & Ref & r=1 & r=10 & r=20 \\
\hline
\hline
SCNCD~\cite{SCNCD} & ECCV2014  & 41.60 & 79.40 & 87.80 \\
Semantic~\cite{Semantic} & CVPR2015 & 44.90 & 77.50 & 86.70 \\
CSL~\cite{CSL} & ICCV2015 & 44.40 &	82.20 & 89.80 \\
XQDA~\cite{XQDA} & CVPR2015 & 61.38 & 91.02 & 95.33 \\
TMA~\cite{TMA} & ECCV2016  & 52.89	& 85.78 & 93.33 \\
LSSCDL~\cite{LSSCDL} & CVPR2016 & 60.49	& 88.58	& 93.60 \\
GOG~\cite{GOG} & CVPR2016 & \textbf{\color{blue}{68.40}} & \textbf{\color{blue}{94.50}} & \textbf{\color{blue}{97.80}} \\
\hline
\hline
SSM (Ours)   &  & \textbf{\color{red}{72.98}} & \textbf{\color{red}{96.76}} & \textbf{\color{red}{99.11}} \\
\hline
\end{tabular}
\caption{The comparison with state-of-the-art on PRID450S.}
\label{table:PRID450S_art}
\vspace{-1ex}
\end{table}

\vspace{1ex}\noindent\textbf{Comparison on CUHK03.}~The comparison on CUHK03 dataset is given in Table~\ref{table:CUHK03_art}. As it shows, the rank-1 identification rate of SSM is $72.7$ with automatically detected bounding boxes, which is the first work reporting rank-1 accuracy larger than $70$.
\begin{table}[tb]
\small
\centering
\begin{tabular}{|l|*{3}{p{0.44cm}<{\centering}}|*{3}{p{0.44cm}<{\centering}}|}
\hline
\multirow{2}{*}{Methods}   & \multicolumn{3}{c|}{Labeled} & \multicolumn{3}{c|}{Detected}\\
\cline{2-7}
          & r=1 & r=5 & r=10  & r=1 & r=5 & r=10      \\
\hline
\hline
DeepReID~\cite{Deepreid} & 20.7 & 51.7 & 68.3 & 19.9 & 49.0 & 64.3 \\
XQDA~\cite{XQDA} & 52.2 & - & - & 46.3 & - & - \\
ImprovedDeep~\cite{ImprovedDeep} & 54.7 & 88.3 & 93.3 & 45.0 & 75.7 & 83.0 \\
LSSCDL~\cite{LSSCDL} & 57.0	& - & - &	51.2 & - & - \\
MLAPG~\cite{MLAPG} & 58.0 & - & - & 51.2 & - & - \\
Shi~\etal~\cite{shi2016embedding} & 61.3	& - & -	& 52.0 & - & - \\
MetricEmsemb.~\cite{MetricEmsemble} & 62.1 & 89.1 & 94.3 & - & - & -\\
Null~\cite{null} & 62.5 & 90.0 & 94.8 &	54.7 &	84.7 &	\textbf{\color{blue}{94.8}} \\
S-LSTM~\cite{S-LSTM} & - & - & - & 57.3 & 80.1 & 88.3 \\
S-CNN~\cite{S-CNN} & - & - & - & 61.8 & 80.9 & 88.3 \\
GOG~\cite{GOG} & 67.3 & \textbf{\color{blue}{91.0}} & \textbf{\color{blue}{96.0}} & \textbf{\color{blue}{65.5}} & \textbf{\color{blue}{88.4}} & 93.7 \\
DGD~\cite{DGD} & \textbf{\color{blue}{75.3}} & - & - & - & - & - \\
\hline
\hline
SSM (Ours) & \textbf{\color{red}{76.6}} & \textbf{\color{red}{94.6}} & \textbf{\color{red}{98.0}} & \textbf{\color{red}{72.7}} & \textbf{\color{red}{92.4}} & \textbf{\color{red}{96.1}}\\
\hline
\end{tabular}
\caption{The comparison with state-of-the-art on CUHK03 dataset.}
\label{table:CUHK03_art}
\vspace{-1ex}
\end{table}

In~\cite{null}, Zhang~\etal~overcome the small sample size (SSS) problem by matching people in a discriminative null space of the training data, which report the second best performance $94.8$ at rank-10 with automatically detected bounding boxes. Nevertheless, the performance gap with SSM becomes larger at lower ranks. For instance, SSM makes a significant improvement of $18.0$ in rank-1 accuracy over Null~\cite{null} with detected bounding boxes.

GOG remains to be one of the most robust descriptors on this dataset. Under XQDA metric, it achieves the second best performances at most ranks. As analyzed above, SSM can be deemed as a generic tool for those visual descriptors and metric learning techniques. Thus, SSM can further enhance their discriminative power.

\subsection{Market-1501}
Market-1501~\cite{market1501} is the largest benchmark in person re-identification up to present, which is comprised of $1501$ identities. $750$ identities ($12,936$ images) are used for training and $751$ identities ($19,732$ images) are used for testing. $3,368$ images are taken as the probe. Both CMC scores and mean average precision (mAP) are used for evaluation.

Thanks to plenty of training images provided, training deep neural networks becomes feasible on this dataset and preferred by most previous works~\cite{SuChi2,S-LSTM,S-CNN}. Following this trend, we introduce Residual Network (ResNet)~\cite{ResNet}, for the first time, to person re-identification.
More specifically, we fine-tune a 50-layer ResNet with classification loss on training images, and extract activations of its last fully connected layer. The $L_2$ normalized activations are taken as visual features and Euclidean metric is utilized to measure the distances between images. The baseline performances are mAP $61.12$ with single query (SQ) and $70.82$ with multiple query (MQ), respectively.

In Table~\ref{table:market1501_art}, we present the experimental comparisons. As can be seen, SSM improves the baseline by mAP $7.68$ for SQ and $5.30$ for MQ. Moreover, SSM outperforms the previous state-of-the-art~\cite{S-CNN} by a very large margin, with the improvement of mAP $29.25$ for SQ and $27.73$ for MQ.
\begin{table}[tb]
\small
\centering
\begin{tabular}{|l|*{2}{p{1.1cm}<{\centering}}|*{2}{p{1.1cm}<{\centering}}|}
\hline
\multirow{2}{*}{Methods}   & \multicolumn{2}{c|}{Single Query} & \multicolumn{2}{c|}{Multiple Query}\\
\cline{2-5}
          & Rank-1 & mAP & Rank-1 & mAP   \\
\hline
\hline
SSDAL~\cite{SuChi2} & 39.40 & 19.60 & 49.00 & 25.80  \\
WARCA~\cite{WARCA} & 45.16 & - & - & - \\
SCSP~\cite{SCSP} & 51.90 & 26.35 & - & - \\
S-LSTM~\cite{S-LSTM} & - & - & 61.60 & 35.31 \\
Null~\cite{null} & 61.02 & 35.68 & 71.56 & 46.03 \\
S-CNN~\cite{S-CNN} & \textbf{\color{blue}{65.88}} & \textbf{\color{blue}{39.55}} & \textbf{\color{blue}{76.04}} & \textbf{\color{blue}{48.45}} \\
\hline
\hline
SSM (Ours) & \textbf{\color{red}{82.21}} & \textbf{\color{red}{68.80}} & \textbf{\color{red}{88.18}} & \textbf{\color{red}{76.18}} \\
\hline
\end{tabular}
\caption{The comparison with state-of-the-art on Market-1501.}
\label{table:market1501_art}
\vspace{-1ex}
\end{table}

\subsection{Time Analysis}
As an additional improvement over metric learning, SSM introduces extra time cost without doubt as analyzed in Sec.~\ref{sec:complexity}. In Table~\ref{table:time}, we present the extra execution time of SSM over XQDA metric.
\begin{table}[tb]
\small
\centering
\begin{tabular}{|l|*{2}{p{1.15cm}<{\centering}}|*{2}{p{1.15cm}<{\centering}}|}
\hline
\multirow{2}{*}{Datasets}        & \multicolumn{2}{c|}{Off-line} & \multicolumn{2}{c|}{On-line}\\
\cline{2-5}
            & \#M & \#A  & \#M  & \#A  \\
\hline
\hline
GRID  & 0.90$s$ & +2.38$s$ & 0.17$s$  &  +10.3$ms$  \\
VIPeR  & 2.19$s$ & +2.22$s$  & 0.19$s$ &  +10.2$ms$  \\
PRID450S  & 1.21$s$ & +0.78$s$ & 0.12$s$ & +3.80$ms$  \\
CUHK03  & 789.6$s$ & +1952$s$ & 0.09$s$ &  +0.516$s$ \\
Market1501  & - & +2769$s$ & 146.11$s$ & +21.68$s$  \\
\hline
\end{tabular}
\caption{\#M denotes the initial time cost of metric learning using XQDA. \#A denotes the extra cost brought by the proposed SSM.}
\label{table:time}
\vspace{-2ex}
\end{table}
As SSM manages transferring the graph-based affinity learning to off-line, the off-line cost is increased especially on larger datasets (\eg,~CUHK03 and Market-1501).
In on-line stage, the extra indexing time brought by SSM only occupies a small percentage on all the datasets except CUHK03. On CUHK03 dataset, indexing using XQDA metric only requires $0.09s$, since CUHK03 has a small gallery. As SSM takes into account the larger training data provided by CUHK03, the extra indexing cost is $0.516s$. Nevertheless, the overall indexing time is still within $1$ second.

\section{Conclusion} \label{sec:con}
In this paper, we do not design robust features or metrics that are superior to others in person re-identification. Instead, we contribute a generic tool called Supervised Smoothed Manifold (SSM), upon which most existing algorithms can easily boost their performances further. SSM is very easy to implement. It can also handle the special kind of labeled data and has potential capacity in large scale ReID.
Comprehensive experiments on five benchmarks demonstrate that SSM not only achieves the best performances, but more importantly, incurs acceptable extra on-line cost.

In the furture, we will investigate how to effectively fuse multiple features~\cite{MetricEmsemble} and apply the proposed SSM to other datasets~\cite{OIM}.


{\small
\bibliographystyle{ieee}
\bibliography{egbib}
}

\end{document}